%% file: sn-article.tex
\documentclass[pdflatex,sn-mathphys-num,sort]{sn-jnl}
\usepackage{graphicx}%
\usepackage{multirow}%
\usepackage{amsmath,amssymb,amsfonts}%
\usepackage{amsthm}%
\usepackage{mathrsfs}%
\usepackage[title]{appendix}%
\usepackage{xcolor}%
\usepackage{textcomp}%
\usepackage{manyfoot}%
\usepackage{booktabs}%
\usepackage{tabularx}
\usepackage{threeparttable}
\usepackage{tikz}
\usetikzlibrary{shapes.geometric, arrows.meta, positioning}
\usepackage{pgfplots}
\pgfplotsset{compat=1.18}
\usepackage{caption}
\usepackage{placeins}
\usepackage[utf8]{inputenc}
\usepackage{hyperref}

\emergencystretch=\maxdimen
\newcommand{\pp}{\,pp}

\begin{document}

\title[LLMs on Iranian Ph.D. ECE Entrance Exams]{Evaluating Large Language Models on Iranian Doctoral Entrance Exams in Electrical and Computer Engineering}

\author[1]{\fnm{Sima} \sur{Akbari Neisiyani}}\email{sima.akbari.neisiyani@gmail.com}
\author*[2]{\fnm{Hossein} \sur{Taghizad}}\email{hossein.taghizad@uqtr.ca}

\affil[1]{\orgname{Independent Researcher}, \orgaddress{\city{Trois-Rivières}, \state{Quebec}, \country{Canada}}}
\affil*[2]{\orgdiv{Department of Electrical and Computer Engineering}, \orgname{Université du Québec à Trois-Rivières}, \orgaddress{\city{Trois-Rivières}, \state{Quebec}, \country{Canada}}}

\abstract{We evaluate four large language models (LLMs) on multiple-choice questions from the 2024 and 2025 Iranian national Ph.D. entrance examinations in Electrical Engineering (EE) and Computer Engineering (CE). Of the 440 questions in the four official booklets, 7 were excluded because the question statement or its figure was ambiguous, leaving 433 questions (186 EE, 247 CE), of which 171 (39.5\%) depend on a figure. Each question was presented in the original Persian and in an English translation, giving 3{,}464 scored responses. GPT-5, Gemini 2.5 Pro, Claude Sonnet 4.5 and Grok 4 were queried zero-shot, with reasoning enabled, through a single commercial web interface (Perplexity) in October 2025, using an automated script. Because every model answers the same items, all inferences use paired tests or question-clustered models. Accuracy separates into two tiers: GPT-5 (79.9\%, 95\% CI 76.6--83.1) and Gemini 2.5 Pro (78.4\%, 74.8--81.9) are statistically indistinguishable, and both outperform Grok 4 (71.7\%) and Claude Sonnet 4.5 (71.3\%) by 6.7--8.7 percentage points (Holm-adjusted $p<0.001$). The English--Persian accuracy difference is small for every model ($-0.2$ to $+2.5$ points; exact McNemar $p\ge0.28$). For three models the 90\% confidence interval lies within $\pm5$ points, so differences larger than this can be excluded. Questions with figures are much harder: the odds ratio is 0.30 (95\% CI 0.21--0.43) after adjusting for discipline. The apparent EE deficit of two models disappears once the larger share of figure-based EE items is accounted for. We describe the dataset, the protocol and its limitations, including the uncontrolled web-retrieval and decoding settings of the interface, and release item-level responses and analysis code.}

\keywords{Large language models, Benchmarking, Multilingual evaluation, Persian, Multimodal reasoning, Electrical and computer engineering, Doctoral entrance examination}

\maketitle

\section{Introduction}
Large Language Models (LLMs) now perform well on a wide range of natural language tasks, including question answering, code generation, translation and scientific reasoning~\cite{minaee_large_2025, matarazzo_survey_2025, ai4science_impact_2023}. As they are increasingly used in tutoring and assessment, it matters how competent they are on demanding, domain-specific academic material~\cite{lothritz_testing_2025, tophel_towards_2025}.

Widely used benchmarks emphasise general knowledge (e.g., MMLU~\cite{hendrycks_measuring_2021}), school mathematics (e.g., GSM8K~\cite{cobbe_training_2021}) or narrower university-level problem sets. Fewer evaluations use authentic, high-stakes graduate examinations in engineering, and fewer still do so in a non-English language with a matched English version that allows a paired language comparison.

This paper reports such an evaluation. We use the 2024 and 2025 doctoral entrance examinations in Electrical and Computer Engineering (ECE), which the Iranian National Organization for Educational Testing (NOET) administers nationwide. The exams cover advanced topics such as electric circuits, electronics, power systems, control, computer architecture, operating systems and machine learning, and many items require reading a circuit diagram, waveform or schematic.

Our contributions are:
\begin{itemize}
    \item \textbf{Dataset.} A bilingual (Persian--English) set of 433 authentic doctoral entrance exam questions. Each item is annotated with discipline, subject area and modality (text-only vs.\ figure-dependent). We document the exclusion of 7 ambiguous items and the translation procedure.
    \item \textbf{Paired evaluation.} Four frontier LLMs (GPT-5, Gemini 2.5 Pro, Claude Sonnet 4.5, Grok 4) answer every item in both languages under a uniform zero-shot protocol. The analysis respects the paired, clustered structure of the data: exact McNemar tests, cluster bootstrap confidence intervals, generalised estimating equations and Holm correction.
    \item \textbf{Findings.} The models fall into two accuracy tiers, and the language effect is small and bounded. Figure-dependent items are the main source of difficulty, and they also explain the apparent EE/CE gap.
    \item \textbf{Transparency.} We report all counts, the model configuration actually used, the scoring procedure and the threats to validity. We release item-level responses and analysis code.
\end{itemize}

\section{Related Work}
Early LLM benchmarks emphasised broad reasoning. MMLU~\cite{hendrycks_measuring_2021} spans many subjects but has limited depth in advanced specialised topics, and GSM8K~\cite{cobbe_training_2021} targets grade-school mathematics. More recent datasets raise the difficulty: university-level multimodal mathematics (U-MATH~\cite{chernyshev_u-math_2025}), Olympiad-level problems (Omni-MATH~\cite{gao_omni-math_2024}) and hierarchical bilingual mathematics (MathBench~\cite{liu_mathbench_2024}).

Multilingual evaluations repeatedly report gaps in non-English settings. MM-Eval~\cite{son_mm-eval_2025} covers 18 languages and exposes weaknesses in low-resource scenarios. Exam-based studies cover Luxembourgish proficiency tests~\cite{lothritz_testing_2025}, Persian computer-science master's entrance questions~\cite{varastehnezhad_llm_2024}, Japanese medical licensing~\cite{liu_performance_2024} and Spanish university admissions~\cite{miranda_evaluating_2023}. They often find lower accuracy for non-English or visually dependent items. The closest prior work is VarastehNezhad et al.~\cite{varastehnezhad_llm_2024}, who studied Iranian master's-level data-structures questions and found most models more accurate in English than in Persian.

Domain-specific exam studies include Hebrew chemistry~\cite{yacobson_benchmarking_2025}, USMLE~\cite{yaneva_examining_2024}, plastic surgery boards~\cite{al_qurashi_can_2023}, veterinary physiology~\cite{bayer_chatgpt-4o_2024}, Brazilian computer-science assessments~\cite{mendonca_evaluating_2024} and hydrology (HydroLLM~\cite{kizilkaya_toward_2025}). They report reasonable performance alongside recurring difficulty with multi-step reasoning and figures. Methodological work warns that benchmark scores need error bars that account for clustered items~\cite{miller_errorbars_2024}, and that training-data contamination must be considered for each benchmark~\cite{sainz_contamination_2023}. Prompting and application studies cover zero-shot prompting~\cite{kojima_large_2022, li_practical_2023}, difficulty estimation~\cite{jain_exploring_2025}, language learning~\cite{zhang_artificial_2025} and clinical support~\cite{wang_accuracy_2025}.

To our knowledge, no previous study evaluates current LLMs on doctoral entrance exams spanning both EE and CE in a paired Persian--English design. We do not claim that the item set is a psychometrically validated benchmark (Section~\ref{sec:limitations}). We present it as a realistic, documented evaluation set.

\section{Dataset}
\subsection{Source exams and item exclusion}
The questions come from the NOET doctoral entrance examinations in EE and CE administered in 2024 and 2025 (Iranian years 1403 and 1404). Each year, the EE booklet contains 95 four-option multiple-choice questions and the CE booklet 125, so the four booklets contain 440 questions in total. The subject-area structure, identical in both years, is given in Table~\ref{tab:exam-structure}. Correct answers were taken from the official NOET answer keys.

Before any model was queried, we reviewed all items and excluded 7 (1.6\%) whose statement or accompanying figure was ambiguous or incomplete in the source booklet, so that a unique correct option could not be established with confidence. The excluded items are EE-2024 Q51, Q52 (Electronics) and Q86 (Linear Control Systems); EE-2025 Q72 (Electrical Energy Systems and Machines); CE-2024 Q45 (Operating Systems); and CE-2025 Q12 (Data Structures and Algorithms) and Q43 (Operating Systems). The final set therefore contains \textbf{433 questions}: 92 EE-2024, 94 EE-2025, 124 CE-2024 and 123 CE-2025. Table~\ref{tab:exam-structure} gives the retained counts per subject and the number of figure-dependent items.

\begin{table}[htbp]
    \centering
    \caption{Exam structure and retained items. ``Official'' is the number of questions per booklet and year; ``Retained'' counts the questions kept for evaluation in 2024 and 2025; ``Fig.'' is the number of retained questions that include a figure.}
    \label{tab:exam-structure}
    \begin{tabularx}{\textwidth}{Xccccc}
        \toprule
        \textbf{Subject area} & \textbf{Official} & \textbf{2024} & \textbf{2025} & \textbf{Total} & \textbf{Fig.} \\
        \midrule
        \multicolumn{6}{l}{\textbf{Electrical Engineering (EE)}} \\
        Engineering Mathematics and Electric Circuits      & 25 & 25 & 25 & 50 & 28 \\
        Signals and Systems                                & 10 & 10 & 10 & 20 & 2  \\
        Electronics                                        & 20 & 18 & 20 & 38 & 38 \\
        Electromagnetics                                   & 10 & 10 & 10 & 20 & 15 \\
        Analysis of Electrical Energy Systems and Machines & 20 & 20 & 19 & 39 & 18 \\
        Linear Control Systems                             & 10 & 9  & 10 & 19 & 12 \\
        \emph{EE subtotal}                                 & 95 & 92 & 94 & 186 & 113 \\
        \midrule
        \multicolumn{6}{l}{\textbf{Computer Engineering (CE)}} \\
        Data Structures and Algorithm Design               & 20 & 20 & 19 & 39 & 3  \\
        Logic Circuits and Computer Architecture           & 20 & 20 & 20 & 40 & 15 \\
        Advanced Operating Systems                         & 15 & 14 & 14 & 28 & 3  \\
        Advanced Databases                                 & 10 & 10 & 10 & 20 & 2  \\
        Pattern Recognition and Machine Learning           & 25 & 25 & 25 & 50 & 11 \\
        Advanced Networks                                  & 10 & 10 & 10 & 20 & 4  \\
        Advanced Computer Architecture and VLSI            & 25 & 25 & 25 & 50 & 20 \\
        \emph{CE subtotal}                                 & 125 & 124 & 123 & 247 & 58 \\
        \midrule
        \textbf{Total}                                     & 220 & 216 & 217 & 433 & 171 \\
        \bottomrule
    \end{tabularx}
\end{table}

\subsection{Translation}
The original questions are in Persian. One author, a bilingual engineer, translated every question into English, and the second author, also bilingual with an engineering background, reviewed each translation against the Persian original; corrections were made before any model was queried. The review focused on technical terminology, quantities and units, and on keeping the four options exactly parallel. Mathematical content was written once in \LaTeX{} and shared between the two versions, so formulas are identical across languages. We did not perform an independent back-translation. Instead, Section~\ref{sec:language} reports an empirical consistency check that flags items whose correctness changes systematically with language.

\subsection{Formatting and annotation}
\begin{figure}[htbp]
    \centering
    \resizebox{0.82\textwidth}{!}{%
        \begin{tikzpicture}[
                node distance=0.55cm,
                box/.style={rectangle, draw, rounded corners, minimum width=24em, minimum height=2.3em, align=center, font=\small},
                startstop/.style={box, fill=green!10},
                process/.style={box, fill=blue!10},
                arrow/.style={thick,->,>=stealth}
            ]
            \node (start) [startstop] {Official NOET Ph.D.\ booklets and answer keys\\EE and CE, 2024 and 2025: 440 questions};
            \node (excl) [process, below=of start] {Item review and exclusion\\7 ambiguous statements or figures removed $\rightarrow$ 433 questions\\(186 EE + 247 CE)};
            \node (translate) [process, below=of excl] {Persian $\to$ English translation by one author,\\reviewed by the second author};
            \node (format) [process, below=of translate] {Prompt formatting: \LaTeX{} math, options 1--4,\\answer-format instruction; figures cropped as PNG};
            \node (annotate) [process, below=of format] {Annotation: discipline, subject area, modality\\(262 text-only, 171 with figure)};
            \node (end) [startstop, below=of annotate] {Evaluation set: 433 items $\times$ 2 languages\\$\times$ 4 models = 3{,}464 scored responses};
            \draw[arrow] (start) -- (excl);
            \draw[arrow] (excl) -- (translate);
            \draw[arrow] (translate) -- (format);
            \draw[arrow] (format) -- (annotate);
            \draw[arrow] (annotate) -- (end);
        \end{tikzpicture}
    }
    \caption{Dataset construction pipeline.}
    \label{fig:dataset-pipeline}
\end{figure}
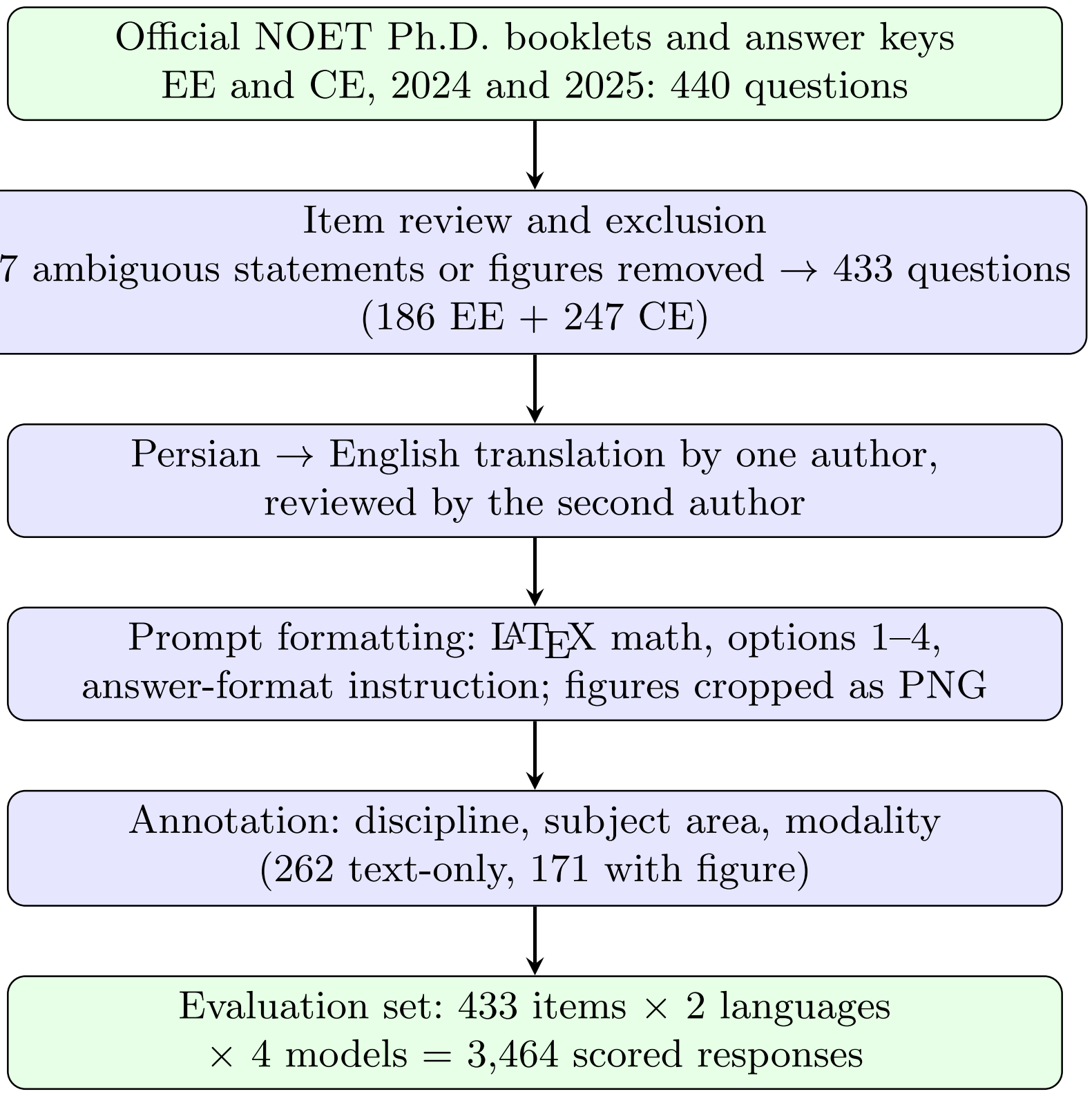

Each question was converted to a plain-text prompt. Mathematical expressions were written in \LaTeX{} with inline \verb|$...$| delimiters. Code fragments were given verbatim, and the four options were listed as a \LaTeX{} \verb|enumerate| list, implicitly numbered 1--4. The prompt ended with an instruction in the language of the question: ``Please answer only with the option number.'' in the English version, and its Persian equivalent in the Persian version. When a question depended on a figure (circuit, waveform, block diagram, schematic or table rendered as an image), only the figure was cropped from the booklet and attached as a PNG file. The question text itself was always given as text. Each item was annotated with its discipline, its subject area (following the official exam structure) and its modality: \emph{text} if it has no figure and \emph{figure} (``image'') otherwise. Figure-dependent items make up 60.8\% of EE items (113/186) but only 23.5\% of CE items (58/247). Every Electronics item has a figure. This imbalance matters for interpreting the discipline comparison (Section~\ref{sec:discipline}). Figure~\ref{fig:dataset-pipeline} summarises the pipeline.

\section{Evaluation Protocol}
\subsection{Models and access}
We evaluated four models that were current at the time of testing: GPT-5 (OpenAI)~\cite{openai_gpt5_2025}, Gemini 2.5 Pro (Google)~\cite{comanici_gemini25_2025}, Claude Sonnet 4.5 (Anthropic)~\cite{anthropic_sonnet45_2025} and Grok 4 (xAI)~\cite{xai_grok4_2025}. All four were accessed through the same web interface, Perplexity~\cite{perplexity_2025}, which lets the user choose the underlying model from a menu. For every model we enabled the interface's ``With reasoning'' option. In the interface this selects ``GPT-5 Thinking'' and ``Claude Sonnet 4.5 Thinking''; Gemini 2.5 Pro and Grok 4 are reasoning models by default. All runs took place in October 2025.

Using one interface means the same submission mechanism, image upload path and timing method apply to all models. It also has consequences that we state explicitly. (i) The interface does not expose model snapshot identifiers, temperature, maximum output length or the system prompt, so these were neither controlled nor recorded; the results describe the models \emph{as deployed in this interface at that time}, not fixed API snapshots. (ii) The interface's default configuration was used, and its web-retrieval feature was not explicitly disabled. Models may therefore have consulted web sources while answering. Section~\ref{sec:limitations} discusses the implications. (iii) Each condition was run once, so run-to-run variability is not measured directly.

\subsection{Procedure}
Submission was automated with a Playwright script that drives a browser session. For each model and language, the script selected the model, verified that reasoning was enabled, and then, for each question, opened a \emph{new thread} so that no conversational context carried over. It attached any figure, typed the prompt and submitted it. No examples, few-shot demonstrations, chain-of-thought instructions or follow-up messages were given (zero-shot)~\cite{li_practical_2023}, and the first response was final. The Persian and English versions were run as separate passes. In total, $433 \times 2 \times 4 = 3{,}464$ responses were collected.

\subsection{Scoring}
The script extracted the visible answer text from each response. One author then read every extracted response and recorded the selected option (1--4), and the recorded option was compared with the official key. Because the prompts asked for the option number only, almost all responses stated a single option unambiguously. In 8 of 3{,}464 responses (0.23\%), concerning 4 CE-2024 items, no single option could be identified; these were scored as incorrect. Scoring was done by a single coder, so we cannot report inter-rater agreement. Section~\ref{sec:limitations} notes this as a limitation.

\subsection{Response time}\label{sec:rt}
Response time is the wall-clock interval from submission until the interface's answer stream completed, as recorded by the script. It therefore includes platform queuing, any web retrieval, network latency and generation, so it should be read as end-to-end latency as experienced by a user, not as model inference time.

\subsection{Statistical analysis}\label{sec:stats}
The 3{,}464 responses are not independent: each question is answered eight times (4 models $\times$ 2 languages), and questions differ greatly in difficulty. All inferences therefore treat the \emph{question} as the unit of resampling or clustering~\cite{miller_errorbars_2024}.
\begin{itemize}
    \item \textbf{Confidence intervals} for accuracies are 95\% percentile intervals from a cluster bootstrap over questions (10{,}000 resamples)~\cite{efron_tibshirani_1993}.
    \item \textbf{Model comparison.} We report an omnibus Wald test from a logistic generalised estimating equation (GEE) with an exchangeable working correlation clustered on question~\cite{liang_zeger_1986}, with model and language as covariates. Cochran's $Q$~\cite{cochran_1950} over the 866 question--language blocks is reported as a non-parametric check. For pairwise comparisons, each question's score for a model is averaged over the two languages, the paired difference between two models is tested across the 433 questions, and the six $p$-values are Holm-adjusted~\cite{holm_1979}.
    \item \textbf{Language comparison.} For each model, English and Persian answers to the same question form a matched pair. We use the exact McNemar test~\cite{mcnemar_1947}, report the paired accuracy difference with Wald 95\% and 90\% CIs, and Holm-adjust across the four models. Because a non-significant test is not evidence of equality, we also apply the two one-sided tests (TOST) logic~\cite{schuirmann_1987} with a pre-specified equivalence margin of $\pm5$ percentage points (pp). Equivalence is concluded when the 90\% CI lies inside the margin. We also report the minimum difference detectable with 80\% power given the observed discordance rate.
    \item \textbf{Discipline, modality and year} vary between questions, not within them. Their effects are estimated with logistic GEEs clustered on question and reported as odds ratios (OR) with robust 95\% CIs. Interactions with model are tested by joint Wald tests. Because figure-dependent items are concentrated in EE, the discipline effect is reported both with and without adjustment for modality.
    \item \textbf{Response times} are skewed and include very long outliers, so we report medians and interquartile ranges without trimming, and compare models with the Friedman test over question--language blocks.
\end{itemize}
Subject-area accuracies (19--50 questions per subject) are reported descriptively. With so few items per subject they carry wide uncertainty (roughly $\pm10$--$15$\pp), and we do not test them. Analyses used Python with \texttt{statsmodels} and \texttt{scipy}.

\section{Results}
\subsection{Overall accuracy}
\begin{figure}[htbp]
    \centering
    \input{fig_accuracy_language.tex}
    \caption{Accuracy by model, overall and by input language. Error bars are 95\% cluster bootstrap CIs over questions ($n=433$ questions per language).}
    \label{fig:accuracy-language}
\end{figure}
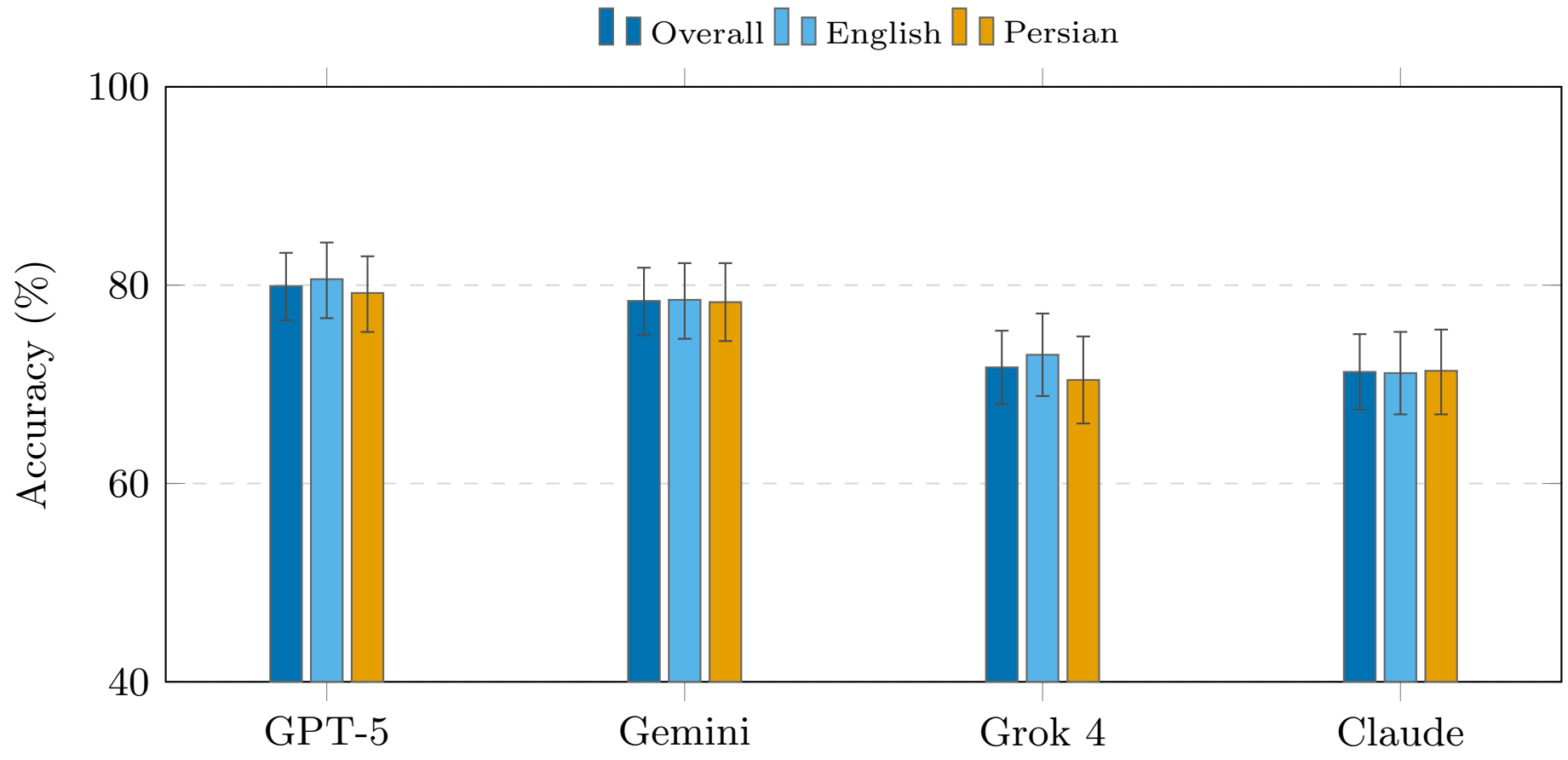

Table~\ref{tab:overall-accuracy} and Figure~\ref{fig:accuracy-language} summarise accuracy over all 866 responses per model (433 questions $\times$ 2 languages). Accuracy differs significantly between models (GEE Wald $\chi^2(3)=38.1$, $p<0.001$; Cochran's $Q(3)=53.4$, $p<0.001$). Pairwise comparisons (Table~\ref{tab:pairwise}) show two tiers. GPT-5 and Gemini 2.5 Pro do not differ significantly (difference $1.5$\pp, 95\% CI $-1.4$ to $4.3$; Holm $p=0.59$), and neither do Grok 4 and Claude Sonnet 4.5 ($0.5$\pp, $-2.8$ to $3.7$; Holm $p=0.78$). Each model in the upper tier is 6.7--8.7\pp{} more accurate than each model in the lower tier (all Holm $p\le0.001$).

\begin{table}[htbp]
    \centering
    \caption{Overall accuracy per model. $k$ is the number of correct responses out of $n$ responses (433 questions $\times$ 2 languages). The 95\% CIs come from a cluster bootstrap over questions.}
    \label{tab:overall-accuracy}
    \begin{tabularx}{0.85\textwidth}{Xccc}
        \toprule
        \textbf{Model}    & $\boldsymbol{k/n}$ & \textbf{Accuracy (\%)} & \textbf{95\% CI} \\
        \midrule
        GPT-5             & 692/866 & 79.91 & 76.56--83.14 \\
        Gemini 2.5 Pro    & 679/866 & 78.41 & 74.83--81.87 \\
        Grok 4            & 621/866 & 71.71 & 67.90--75.40 \\
        Claude Sonnet 4.5 & 617/866 & 71.25 & 67.44--74.94 \\
        \bottomrule
    \end{tabularx}
\end{table}

\begin{table}[htbp]
    \centering
    \caption{Pairwise model differences in accuracy (row model minus column model, percentage points). Each difference is computed on per-question scores averaged over the two languages ($n=433$ questions). The 95\% CIs come from a cluster bootstrap, and $p$-values are Holm-adjusted over the six comparisons.}
    \label{tab:pairwise}
    \begin{tabularx}{\textwidth}{Xccc}
        \toprule
        \textbf{Comparison} & \textbf{Difference (pp)} & \textbf{95\% CI} & \textbf{Holm $p$} \\
        \midrule
        GPT-5 $-$ Gemini 2.5 Pro             & $+1.50$ & $-1.39$ to $4.27$  & 0.594 \\
        GPT-5 $-$ Grok 4                     & $+8.20$ & $5.08$ to $11.43$  & $<0.001$ \\
        GPT-5 $-$ Claude Sonnet 4.5          & $+8.66$ & $5.20$ to $12.01$  & $<0.001$ \\
        Gemini 2.5 Pro $-$ Grok 4            & $+6.70$ & $3.46$ to $10.05$  & $<0.001$ \\
        Gemini 2.5 Pro $-$ Claude Sonnet 4.5 & $+7.16$ & $3.93$ to $10.28$  & $<0.001$ \\
        Grok 4 $-$ Claude Sonnet 4.5         & $+0.46$ & $-2.77$ to $3.70$  & 0.784 \\
        \bottomrule
    \end{tabularx}
\end{table}

\subsection{Persian versus English}\label{sec:language}
Table~\ref{tab:language} gives the paired language comparison for each model. The English-minus-Persian difference ranges from $-0.2$\pp{} (Claude Sonnet 4.5) to $+2.5$\pp{} (Grok 4). No difference is significant (exact McNemar $p\ge0.28$ before correction; all Holm-adjusted $p=1.00$). The pooled GEE estimate across models is also null (OR Persian vs.\ English $=0.95$, 95\% CI 0.85--1.06, $p=0.35$).

Beyond the absence of significance, the data bound the size of any language effect. For GPT-5, Gemini 2.5 Pro and Claude Sonnet 4.5, the 90\% CI of the difference lies entirely within $\pm5$\pp, so by the TOST criterion their accuracies are equivalent across languages at that margin. For Grok 4 the 90\% CI reaches $+6.0$\pp, so a moderate English advantage cannot be ruled out. Given the observed discordance rates, the design has 80\% power to detect differences of about 5--6\pp; smaller language effects cannot be excluded.

The matched design also shows that the null aggregate effect hides substantial item-level instability. For every model, 13.6--19.6\% of questions are answered correctly in one language and incorrectly in the other, and the same option is chosen in both languages for only 74.8--83.4\% of questions. These flips are roughly balanced in direction, so they cancel in the accuracy totals. As a check on translation, we searched for items whose correctness changed in the same direction for at least three of the four models, which would indicate a systematic difference between the two versions. Only 5 of 433 items meet this criterion, and only 1 changes for all four models. We take this as indirect evidence that the translation did not systematically alter item content, while acknowledging that it does not replace formal back-translation.

\begin{table}[htbp]
    \centering
    \caption{Paired Persian--English comparison ($n=433$ questions per model). EN, FA: accuracy (\%) in English and Persian. EN only / FA only: number of questions answered correctly only in English / only in Persian. $\Delta$: English minus Persian accuracy (pp), with its 95\% CI. $p$: exact McNemar test (all Holm-adjusted $p=1.00$). Equiv.: 90\% CI within $\pm5$\pp. Same: percentage of questions for which the same option was chosen in both languages.}
    \label{tab:language}
    \footnotesize
    \setlength{\tabcolsep}{3.5pt}
    \begin{tabular*}{\textwidth}{@{\extracolsep{\fill}}lccccrcccc@{}}
        \toprule
        \textbf{Model} & \textbf{EN} & \textbf{FA} & \textbf{EN only} & \textbf{FA only} & $\boldsymbol{\Delta}$ & \textbf{95\% CI} & $\boldsymbol{p}$ & \textbf{Equiv.} & \textbf{Same} \\
        \midrule
        GPT-5             & 80.60 & 79.21 & 34 & 28 & $+1.39$ & $-2.18$, $4.95$ & 0.53 & yes & 83.4 \\
        Gemini 2.5 Pro    & 78.52 & 78.29 & 30 & 29 & $+0.23$ & $-3.25$, $3.71$ & 1.00 & yes & 82.7 \\
        Grok 4            & 72.98 & 70.44 & 48 & 37 & $+2.54$ & $-1.63$, $6.71$ & 0.28 & no  & 74.8 \\
        Claude Sonnet 4.5 & 71.13 & 71.36 & 37 & 38 & $-0.23$ & $-4.15$, $3.69$ & 1.00 & yes & 74.8 \\
        \bottomrule
    \end{tabular*}
\end{table}

\subsection{Discipline and modality}\label{sec:discipline}
Table~\ref{tab:disc-modality} reports accuracy by discipline and modality, with the number of questions in each cell. Comparing disciplines directly, GPT-5 and Gemini 2.5 Pro are slightly \emph{more} accurate on EE than on CE ($+1.8$ and $+1.1$\pp), whereas Grok 4 and Claude Sonnet 4.5 are markedly less accurate on EE ($-7.9$ and $-13.2$\pp). Accordingly, the model $\times$ discipline interaction is significant (Wald $\chi^2(3)=21.0$, $p<0.001$), and the pooled EE main effect is modest (OR $0.70$, 95\% CI 0.51--0.96, $p=0.028$).

\begin{figure}[htbp]
    \centering
    \input{fig_disc_modality.tex}
    \caption{Accuracy by discipline and modality. Error bars are 95\% cluster bootstrap CIs over questions. Questions per cell: CE text 189, CE figure 58, EE text 73, EE figure 113.}
    \label{fig:disc-modality}
\end{figure}
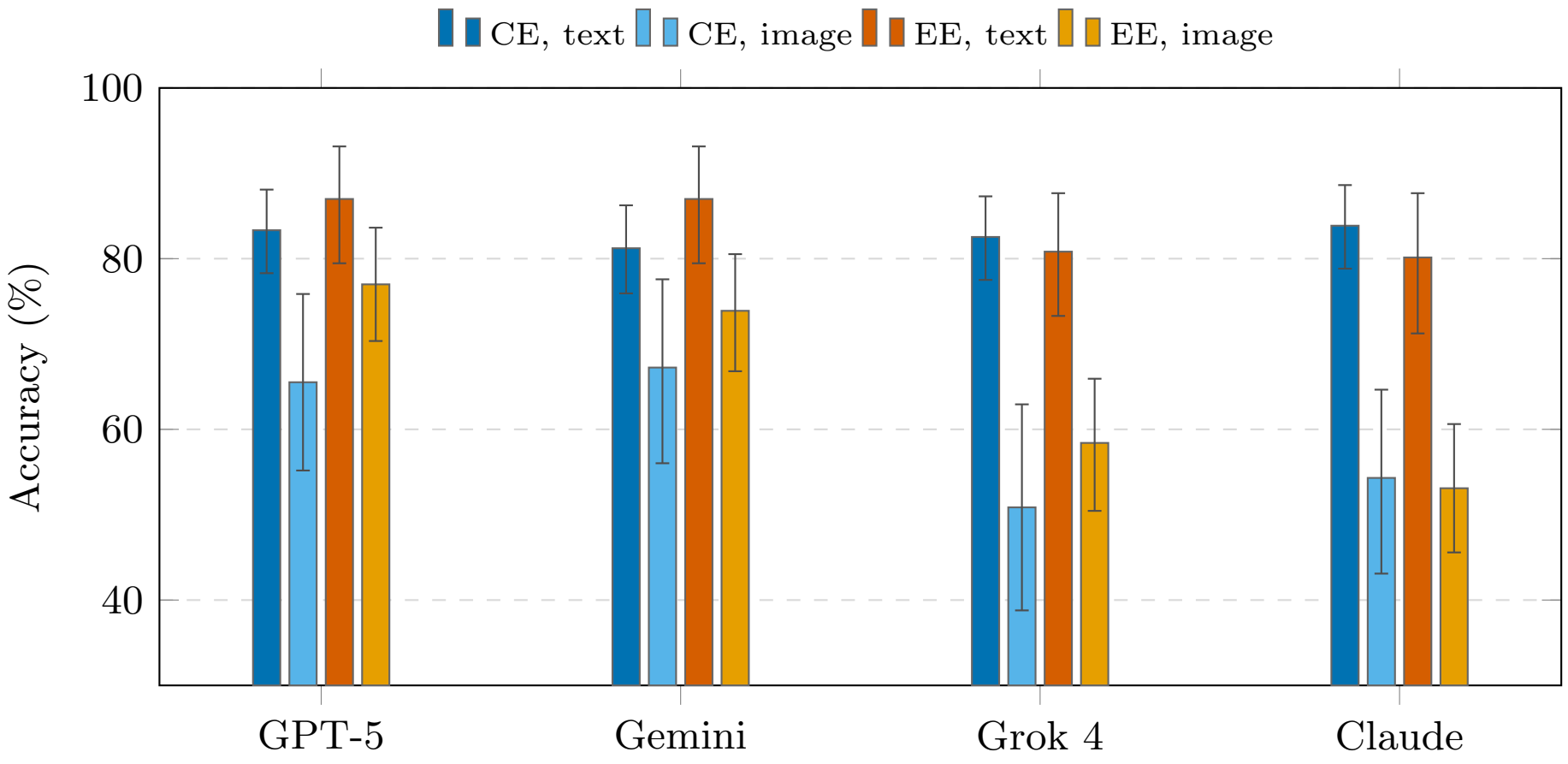

Modality has a much larger effect. Every model is less accurate on questions with a figure, by 11.2\pp{} (GPT-5) to 29.3\pp{} (Claude Sonnet 4.5) (Figure~\ref{fig:disc-modality}). Adjusting for discipline, a figure lowers the odds of a correct answer to 0.30 times those for a text-only question (95\% CI 0.21--0.43, $p<0.001$). The penalty differs significantly between models (model $\times$ modality Wald $\chi^2(3)=26.7$, $p<0.001$): it is smaller for GPT-5 and Gemini 2.5 Pro than for Grok 4 and Claude Sonnet 4.5. Once modality is in the model, the EE effect disappears (OR $1.10$, 95\% CI 0.77--1.56, $p=0.61$). On text-only items, all four models reach 80--87\% in both disciplines (Table~\ref{tab:disc-modality}). The apparent EE weakness of Grok 4 and Claude Sonnet 4.5 is therefore better explained by EE's large share of figure-dependent questions than by EE subject matter as such.

\begin{table}[htbp]
    \centering
    \caption{Accuracy (\%) by discipline and modality. $n_q$ is the number of questions; each model contributes $2n_q$ responses per cell.}
    \label{tab:disc-modality}
    \begin{tabularx}{\textwidth}{Xcccccc}
        \toprule
         & \multicolumn{2}{c}{\textbf{Discipline}} & \multicolumn{2}{c}{\textbf{Modality}} & \multicolumn{2}{c}{\textbf{EE only}} \\
        \cmidrule(lr){2-3}\cmidrule(lr){4-5}\cmidrule(lr){6-7}
        \textbf{Model} & CE & EE & Text & Figure & Text & Figure \\
         & {\footnotesize $n_q=247$} & {\footnotesize $n_q=186$} & {\footnotesize $n_q=262$} & {\footnotesize $n_q=171$} & {\footnotesize $n_q=73$} & {\footnotesize $n_q=113$} \\
        \midrule
        GPT-5             & 79.15 & 80.91 & 84.35 & 73.10 & 86.99 & 76.99 \\
        Gemini 2.5 Pro    & 77.94 & 79.03 & 82.82 & 71.64 & 86.99 & 73.89 \\
        Grok 4            & 75.10 & 67.20 & 82.06 & 55.85 & 80.82 & 58.41 \\
        Claude Sonnet 4.5 & 76.92 & 63.71 & 82.82 & 53.51 & 80.14 & 53.10 \\
        \bottomrule
    \end{tabularx}
\end{table}

\subsection{Subject areas}
Table~\ref{tab:subject-accuracy} gives accuracy per subject area together with the number of questions and figure-dependent items. These results are descriptive (Section~\ref{sec:stats}). The lowest accuracies occur in Electronics, where every item has a figure; Grok 4 and Claude Sonnet 4.5 score 57.9\% and 52.6\% there. Subjects with mostly textual items, such as Data Structures and Algorithm Design (3/39 with figures), Advanced Databases and Pattern Recognition and Machine Learning, yield 80--87\% for all models. This is consistent with the modality effect above.

\begin{table}[htbp]
    \centering
    \caption{Accuracy (\%) by subject area. $n_q$: questions retained; Fig.: of which with a figure. Each value is based on $2n_q$ responses (both languages). With $n_q\le50$, differences of less than about 10--15\pp{} between cells should not be interpreted.}
    \label{tab:subject-accuracy}
    \small
    \begin{tabularx}{\textwidth}{Xcccccc}
        \toprule
        \textbf{Subject area} & $\boldsymbol{n_q}$ & \textbf{Fig.} & \textbf{GPT-5} & \textbf{Gemini} & \textbf{Grok} & \textbf{Claude} \\
        \midrule
        \multicolumn{7}{l}{\textbf{EE}} \\
        Engineering Mathematics and Electric Circuits      & 50 & 28 & 81.00 & 85.00 & 73.00 & 65.00 \\
        Signals and Systems                                & 20 & 2  & 87.50 & 82.50 & 75.00 & 75.00 \\
        Electronics                                        & 38 & 38 & 72.37 & 68.42 & 57.89 & 52.63 \\
        Electromagnetics                                   & 20 & 15 & 87.50 & 82.50 & 72.50 & 62.50 \\
        Analysis of Electrical Energy Systems and Machines & 39 & 18 & 80.77 & 78.21 & 65.38 & 67.95 \\
        Linear Control Systems                             & 19 & 12 & 84.21 & 78.95 & 60.53 & 63.16 \\
        \midrule
        \multicolumn{7}{l}{\textbf{CE}} \\
        Data Structures and Algorithm Design               & 39 & 3  & 87.18 & 87.18 & 85.90 & 80.77 \\
        Logic Circuits and Computer Architecture           & 40 & 15 & 73.75 & 73.75 & 70.00 & 72.50 \\
        Advanced Operating Systems                         & 28 & 3  & 83.93 & 71.43 & 67.86 & 69.64 \\
        Advanced Databases                                 & 20 & 2  & 85.00 & 82.50 & 80.00 & 82.50 \\
        Pattern Recognition and Machine Learning           & 50 & 11 & 85.00 & 82.00 & 84.00 & 83.00 \\
        Advanced Networks                                  & 20 & 4  & 67.50 & 80.00 & 67.50 & 75.00 \\
        Advanced Computer Architecture and VLSI            & 50 & 20 & 71.00 & 71.00 & 67.00 & 74.00 \\
        \bottomrule
    \end{tabularx}
\end{table}

\subsection{Item-level characteristics and contamination probe}
\begin{figure}[htbp]
    \centering
    \input{fig_item_difficulty.tex}
    \caption{Distribution of questions by the number of correct runs out of eight (four models, two languages; $n=433$ questions).}
    \label{fig:items}
\end{figure}
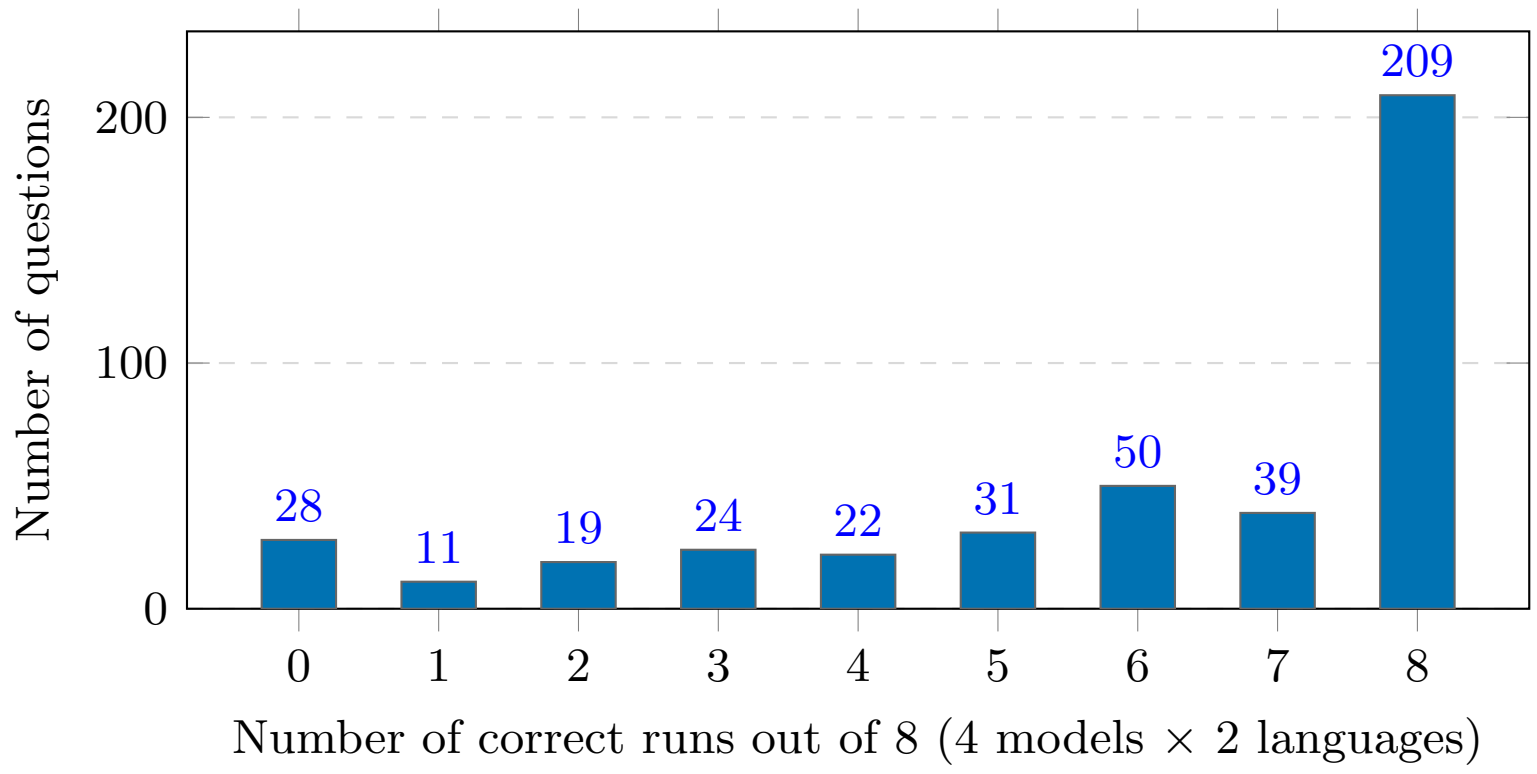

Figure~\ref{fig:items} shows, for each question, how many of the eight model--language runs answered it correctly. The distribution is strongly skewed: 209 questions (48.3\%) are answered correctly in all eight runs, and 28 (6.5\%) in none. The mean proportion correct per item is 0.75. Agreement between models on which items they get right is moderate (Fleiss' $\kappa=0.47$ across the four models~\cite{fleiss_1971}). Thus nearly half the items no longer separate these models, and the comparisons above rest mainly on the remaining half.

The 2024 and 2025 exams were both publicly released before testing. Accuracy does not differ between years after adjusting for model, language, discipline and modality (OR 2025 vs.\ 2024 $=0.94$, 95\% CI 0.68--1.30, $p=0.73$). This probe is weak: both years may be equally exposed, through training data or through web retrieval at test time.

\subsection{Response time}
\begin{figure}[htbp]
    \centering
    \input{fig_response_time.tex}
    \caption{Median end-to-end response time by model and modality. Error bars span the interquartile range.}
    \label{fig:response-time}
\end{figure}
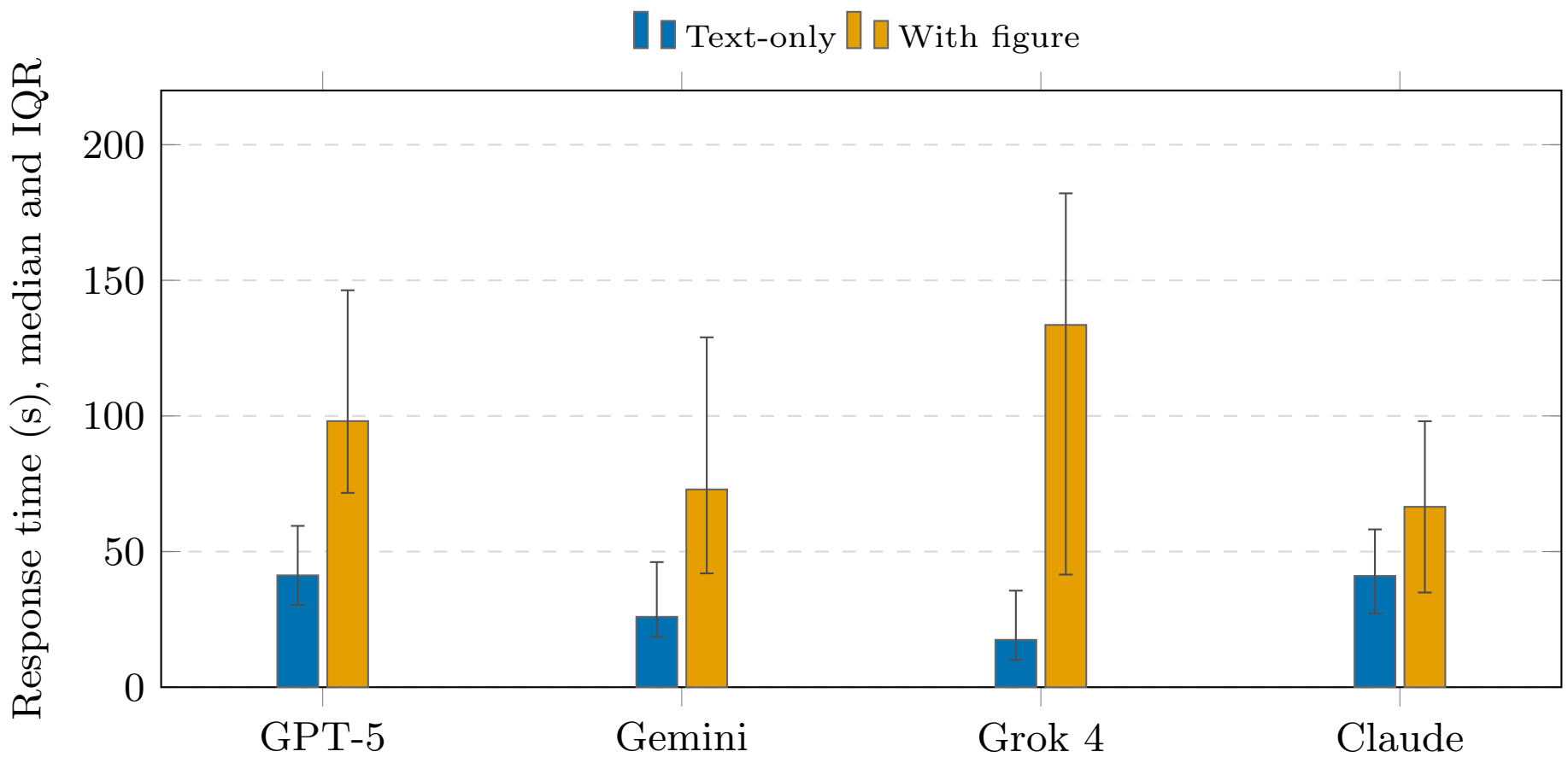

Table~\ref{tab:response_time} and Figure~\ref{fig:response-time} report end-to-end response times over all 866 responses per model, without trimming. Latency differs significantly between models (Friedman $\chi^2(3)=184.3$, $p<0.001$, 866 blocks). The models' rankings depend on the summary statistic. By median, Grok 4 is the fastest (32.0~s) and GPT-5 the slowest (57.8~s). By mean, Claude Sonnet 4.5 is the fastest (57.6~s), because Grok 4's distribution has a long tail (upper quartile 130.6~s; maximum 508~s). Questions with figures take longer for every model (Holm-adjusted Mann--Whitney $p<0.001$ for each). The increase is most pronounced for Grok 4 (median 17.4~s text vs.\ 133.6~s figure). Incorrect answers take longer than correct ones for every model, suggesting that latency partly reflects item difficulty. Median times for Persian and English prompts differ by less than 10~s for every model. Because the timings include platform and retrieval overhead (Section~\ref{sec:rt}), they compare the four configurations as experienced through this interface, not intrinsic inference speed.

\begin{table}[htbp]
    \centering
    \caption{End-to-end response time in seconds (866 responses per model, untrimmed). Med: median; IQR: interquartile range.}
    \label{tab:response_time}
    \footnotesize
    \setlength{\tabcolsep}{4pt}
    \begin{tabular*}{\textwidth}{@{\extracolsep{\fill}}lccccccc@{}}
        \toprule
        \textbf{Model} & \textbf{Med} & \textbf{IQR} & \textbf{Mean} & \textbf{Med text} & \textbf{Med figure} & \textbf{Med correct} & \textbf{Med wrong} \\
        \midrule
        GPT-5             & 57.8 & 36.5--96.2  & 77.5 & 41.2 & 98.1  & 51.7 & 89.1 \\
        Gemini 2.5 Pro    & 38.4 & 22.0--81.8  & 62.8 & 25.9 & 72.9  & 35.2 & 59.7 \\
        Grok 4            & 32.0 & 13.9--130.6 & 76.0 & 17.4 & 133.6 & 24.3 & 94.3 \\
        Claude Sonnet 4.5 & 45.3 & 29.1--78.0  & 57.6 & 41.0 & 66.5  & 42.2 & 67.8 \\
        \bottomrule
    \end{tabular*}
\end{table}

\section{Discussion}
\paragraph{Model tiers.} On authentic doctoral-level ECE questions, the two strongest configurations, GPT-5 and Gemini 2.5 Pro, answered about four in five questions correctly and could not be separated statistically. Grok 4 and Claude Sonnet 4.5 formed a second tier about 7--9\pp{} lower. Given the single-run design and the interface-level configuration, these results rank the four systems as accessed through this interface in October 2025. They are not intrinsic properties of the underlying models.

\paragraph{Language.} Unlike earlier reports of substantial non-English penalties~\cite{son_mm-eval_2025}, including on Iranian master's-level questions with earlier models~\cite{varastehnezhad_llm_2024}, we find no detectable aggregate difference between Persian and English. For three of four models we can bound the difference within $\pm5$\pp. Plausible contributors are the identical \LaTeX{} rendering of mathematics in both versions, the technical vocabulary shared across languages, and improvements in multilingual training in recent models. The aggregate equivalence coexists with considerable item-level inconsistency: up to a fifth of items flip between correct and incorrect across languages. A user asking the same question in a different language may well receive a different answer, even though accuracy over the whole set is unchanged.

\paragraph{Figures, not discipline, drive difficulty.} The clearest result is the penalty for figure-dependent questions, which is large for every model and particularly large for Grok 4 and Claude Sonnet 4.5. Because EE contains 2.6 times the proportion of figure-dependent items found in CE, comparing disciplines without accounting for modality would wrongly suggest that these models are weak at EE content. On text-only questions the discipline gap essentially disappears. This agrees with reports that diagram interpretation remains a weakness of current multimodal models~\cite{chernyshev_u-math_2025}. It also shows that subject-level comparisons in exam benchmarks should account for item format.

\paragraph{Implications.} For educational use, such as tutoring or explaining exam solutions, the strongest models are already accurate on text-only doctoral-level ECE questions. Their reliability drops markedly when a circuit or schematic must be read, and answers to individual items are not stable across languages. These observations argue for human verification, especially for figure-based material. Nearly half the items were solved by all runs, so future versions of this evaluation set will need harder or more discriminating items to keep separating models.

\paragraph{Limitations.}\label{sec:limitations}
\begin{enumerate}
    \item \textbf{Interface, not API.} Models were accessed through a third-party web interface whose snapshot identifiers, system prompt, sampling temperature and output limits are not visible and may change without notice. Exact replication is therefore not possible. We mitigate this by releasing every recorded answer, the prompts' format and the submission script, and we encourage replication through versioned APIs.
    \item \textbf{Web retrieval and contamination.} The interface's web retrieval was not explicitly disabled, and the exams and keys were public. Correct answers may therefore partly reflect retrieval or memorisation rather than reasoning. The absence of a year effect does not rule this out, and absolute accuracies should be read as upper bounds on closed-book performance.
    \item \textbf{Single run.} Each condition was run once, so run-to-run variance was not estimated. The cross-language option agreement of 75--83\% suggests that answer instability is not negligible.
    \item \textbf{Single coder.} Options were recorded by one author. The answer-format instruction made extraction largely mechanical, but no independent double coding or agreement statistic is available.
    \item \textbf{Translation.} Translations were produced by one author and reviewed by the other, without independent back-translation. The cross-model consistency check flags few systematically affected items, but it cannot detect errors that leave correctness unchanged.
    \item \textbf{Item set.} The 433 items come from one country's exams in two years and were not psychometrically calibrated on human examinees. No human score distribution was available, so we make no claims about human-level performance. Subject-level results rest on 19--50 items each. Seven ambiguous items were excluded, which may slightly favour the models.
    \item \textbf{Prompting.} Only zero-shot prompting with reasoning enabled was evaluated. Other prompting strategies~\cite{kojima_large_2022} or tool use might change the results.
\end{enumerate}

\paragraph{Future work.} Natural extensions are replication through versioned APIs with fixed decoding settings and retrieval disabled, repeated runs to estimate stochastic variability, double-coded scoring, back-translation, inclusion of human examinee statistics where obtainable, and a qualitative error analysis of figure-based failures.

\section{Conclusion}
We evaluated GPT-5, Gemini 2.5 Pro, Claude Sonnet 4.5 and Grok 4 on 433 authentic questions from the 2024 and 2025 Iranian Ph.D.\ entrance exams in Electrical and Computer Engineering, in Persian and English. Using paired, question-clustered analyses, we find two accuracy tiers: about 79--80\% for GPT-5 and Gemini 2.5 Pro and about 71--72\% for Grok 4 and Claude Sonnet 4.5. Language effects are small, and within $\pm5$\pp{} for three models. The penalty for questions with figures is large, and it, rather than the electrical-engineering subject matter itself, accounts for the apparent EE weakness of some models. We release the item-level data and code, and we document the limitations of interface-based testing, so that these results can be checked and extended.

\section*{Data and Code Availability}
The item metadata (identifier, year, discipline, subject area, modality), the official answer keys, every recorded model answer and response time, and the analysis code that reproduces all tables and figures are provided as ancillary files with this preprint. The original question booklets are published by NOET. The English translations and the cropped figures are available from the corresponding author on reasonable request.

\section*{Ethical Considerations}
This study involves no human participants or personal data. All examination questions are publicly released NOET materials.

\section*{Competing Interests}
The authors declare no competing financial interests or personal relationships that could have appeared to influence the work reported in this paper.

\section*{Funding}
This research received no specific grant from any funding agency in the public, commercial, or not-for-profit sectors.

\section*{Acknowledgements}
The evaluation set was constructed from publicly available Ph.D.\ entrance examination materials administered by the National Organization for Educational Testing (NOET), Iran. The authors acknowledge NOET as the original source of these materials.

\FloatBarrier
\bibliography{sn-bibliography}
\end{document}

%% file: fig_accuracy_language.tex
\begin{tikzpicture}
\begin{axis}[
  width=0.95\textwidth, height=6.2cm, ybar, bar width=7pt,
  ymin=40, ymax=100, ylabel={Accuracy (\%)},
  symbolic x coords={GPT-5,Gemini,Grok 4,Claude}, xtick=data,
  enlarge x limits=0.15, ymajorgrids, grid style={dashed,gray!30},
  tick label style={font=\small}, label style={font=\small},
  legend style={at={(0.5,1.03)},anchor=south,legend columns=3,font=\footnotesize,draw=none},
  error bars/y dir=both, error bars/y explicit,
  error bars/error bar style={black!70}, error bars/error mark options={rotate=90,mark size=1.5pt,black!70},
]
\addplot+[draw=black!60, fill={rgb,255:red,0;green,114;blue,178}] coordinates {(GPT-5,79.91) += (0,3.35) -= (0,3.46) (Gemini,78.41) += (0,3.35) -= (0,3.46) (Grok 4,71.71) += (0,3.70) -= (0,3.70) (Claude,71.25) += (0,3.81) -= (0,3.81)};
\addlegendentry{Overall}
\addplot+[draw=black!60, fill={rgb,255:red,86;green,180;blue,233}] coordinates {(GPT-5,80.60) += (0,3.70) -= (0,3.93) (Gemini,78.52) += (0,3.70) -= (0,3.93) (Grok 4,72.98) += (0,4.16) -= (0,4.16) (Claude,71.13) += (0,4.16) -= (0,4.16)};
\addlegendentry{English}
\addplot+[draw=black!60, fill={rgb,255:red,230;green,159;blue,0}] coordinates {(GPT-5,79.21) += (0,3.70) -= (0,3.93) (Gemini,78.29) += (0,3.93) -= (0,3.93) (Grok 4,70.44) += (0,4.39) -= (0,4.39) (Claude,71.36) += (0,4.16) -= (0,4.39)};
\addlegendentry{Persian}
\end{axis}
\end{tikzpicture}

%% file: fig_disc_modality.tex
\begin{tikzpicture}
\begin{axis}[
  width=0.95\textwidth, height=6.2cm, ybar, bar width=6pt,
  ymin=30, ymax=100, ylabel={Accuracy (\%)},
  symbolic x coords={GPT-5,Gemini,Grok 4,Claude}, xtick=data,
  enlarge x limits=0.15, ymajorgrids, grid style={dashed,gray!30},
  tick label style={font=\small}, label style={font=\small},
  legend style={at={(0.5,1.03)},anchor=south,legend columns=4,font=\footnotesize,draw=none},
  error bars/y dir=both, error bars/y explicit,
  error bars/error bar style={black!70}, error bars/error mark options={rotate=90,mark size=1.5pt,black!70},
]
\addplot+[draw=black!60, fill={rgb,255:red,0;green,114;blue,178}] coordinates {(GPT-5,83.33) += (0,4.76) -= (0,5.03) (Gemini,81.22) += (0,5.03) -= (0,5.29) (Grok 4,82.54) += (0,4.76) -= (0,5.03) (Claude,83.86) += (0,4.76) -= (0,5.03)};
\addlegendentry{CE, text}
\addplot+[draw=black!60, fill={rgb,255:red,86;green,180;blue,233}] coordinates {(GPT-5,65.52) += (0,10.34) -= (0,10.34) (Gemini,67.24) += (0,10.34) -= (0,11.21) (Grok 4,50.86) += (0,12.07) -= (0,12.07) (Claude,54.31) += (0,10.34) -= (0,11.21)};
\addlegendentry{CE, image}
\addplot+[draw=black!60, fill={rgb,255:red,213;green,94;blue,0}] coordinates {(GPT-5,86.99) += (0,6.16) -= (0,7.53) (Gemini,86.99) += (0,6.16) -= (0,7.53) (Grok 4,80.82) += (0,6.85) -= (0,7.53) (Claude,80.14) += (0,7.53) -= (0,8.90)};
\addlegendentry{EE, text}
\addplot+[draw=black!60, fill={rgb,255:red,230;green,159;blue,0}] coordinates {(GPT-5,76.99) += (0,6.64) -= (0,6.64) (Gemini,73.89) += (0,6.64) -= (0,7.08) (Grok 4,58.41) += (0,7.52) -= (0,7.96) (Claude,53.10) += (0,7.52) -= (0,7.52)};
\addlegendentry{EE, image}
\end{axis}
\end{tikzpicture}

%% file: fig_item_difficulty.tex
\begin{tikzpicture}
\begin{axis}[width=0.8\textwidth, height=5.4cm, ybar, bar width=14pt,
  xlabel={Number of correct runs out of 8 (4 models $\times$ 2 languages)}, ylabel={Number of questions},
  xtick={0,...,8}, ymin=0, ymax=235, ymajorgrids, grid style={dashed,gray!30},
  nodes near coords, nodes near coords style={font=\scriptsize},
  tick label style={font=\small}, label style={font=\small}]
\addplot+[draw=black!60, fill={rgb,255:red,0;green,114;blue,178}] coordinates {(0,28) (1,11) (2,19) (3,24) (4,22) (5,31) (6,50) (7,39) (8,209)};
\end{axis}
\end{tikzpicture}

%% file: fig_response_time.tex
\begin{tikzpicture}
\begin{axis}[
  width=0.95\textwidth, height=6.2cm, ybar, bar width=9pt,
  ymin=0, ymax=220, ylabel={Response time (s), median and IQR},
  symbolic x coords={GPT-5,Gemini,Grok 4,Claude}, xtick=data,
  enlarge x limits=0.15, ymajorgrids, grid style={dashed,gray!30},
  tick label style={font=\small}, label style={font=\small},
  legend style={at={(0.5,1.03)},anchor=south,legend columns=2,font=\footnotesize,draw=none},
  error bars/y dir=both, error bars/y explicit,
  error bars/error bar style={black!70}, error bars/error mark options={rotate=90,mark size=1.5pt,black!70},
]
\addplot+[draw=black!60, fill={rgb,255:red,0;green,114;blue,178}] coordinates {(GPT-5,41.22) += (0,18.25) -= (0,10.92) (Gemini,25.94) += (0,20.18) -= (0,7.35) (Grok 4,17.45) += (0,18.14) -= (0,7.35) (Claude,41.02) += (0,17.16) -= (0,13.80)};
\addlegendentry{Text-only}
\addplot+[draw=black!60, fill={rgb,255:red,230;green,159;blue,0}] coordinates {(GPT-5,98.09) += (0,48.25) -= (0,26.46) (Gemini,72.90) += (0,56.09) -= (0,30.94) (Grok 4,133.56) += (0,48.50) -= (0,92.05) (Claude,66.50) += (0,31.55) -= (0,31.60)};
\addlegendentry{With figure}
\end{axis}
\end{tikzpicture}